\pdfoutput=1

\documentclass[table, 11pt]{article}

\usepackage{acl}

\usepackage{times}
\usepackage{latexsym}
\usepackage{todonotes}

\usepackage{soul}
\usepackage{mathtools}
\usepackage{caption}
\usepackage{cleveref}

\usepackage{subcaption}
\usepackage{microtype}
\usepackage{booktabs}
\usepackage{amsfonts}
\usepackage{multicol, multirow}
\usepackage{xcolor}
\usepackage{graphbox} 
\usepackage{nicematrix}
\usepackage[linesnumbered,ruled,vlined]{algorithm2e}

\usepackage[T1]{fontenc}

\usepackage[utf8]{inputenc}

\newcommand{\methodname}{\textsc{InstrExp}}
\newcommand{\origdatasetname}{\textsc{MIns}}
\newcommand{\datasetname}{\textsc{MIns+}}
\newcommand{\datasetmtname}{$\textsc{MIns+}_\textrm{MT}$}
\newcommand{\datasetitername}{$\textsc{MIns+}_\textrm{Iter}$}

\newcommand{\inc}[1]{{\footnotesize\color{red}(+#1)}}
\newcommand{\dec}[1]{{\footnotesize\color{blue}(-#1)}}
\definecolor{Gray}{gray}{0.9}

\Crefname{algocf}{Algorithm}{Algorithms}

\title{Towards Robust Instruction Tuning on Multimodal Large Language Models}

\author{
    Wei Han \textsuperscript{1} \quad
    Hui Chen \textsuperscript{2} \quad
    Soujanya Poria \textsuperscript{1} \\
    \textsuperscript{1} Singapore University of Technology and Design, \textsuperscript{2} Nanyang Technological University \\
}

\begin{document}
\maketitle

\begin{abstract}
Fine-tuning large language models (LLMs) on instruction-following data across multiple tasks has been proven an effective learning paradigm for enhancing their zero-shot performance on downstream tasks specified through human language. The success has also been extended on multimodal large language models (MLLMs).
Despite the effectiveness, these fine-tuned models often exhibit weak robustness when encountering instructions writing in various wording styles.
In this work, we present an automatic instruction set expansion  method~\methodname, targeted on compositional instruction template consists of both placeholders and natural text. 
Starting from a handful of basic and straightforward meta instructions but can expand an instruction-following dataset by 30 times.
Results on three multimodal instruction-following benchmarks and three MLLMs show that~\methodname~can significantly improve the alignment of multimodal large language models (MLLMs), which is even equivalent to the benefits of scaling up training data multiple times. 
\end{abstract}

\section{Introduction}
In recent years we have witnessed a surge in the family of large language models (LLMs). 
Thanks to the scaling laws for LLMs~\cite{chowdhery2023palm}, a series of works have established powerful foundation models which show superior zero-shot performance on the downstream tasks~\cite{brown2020language,touvron2023llama,touvron2023llama2,openai2023gpt,team2023gemini,bai2023qwen}.
Instruction fine-tuning (IFT), which aims to ``teach" models to follow natural language instructions, has been proven to be an effective learning paradigm to further enhance LLMs' generalizability~\cite{sanh2021multitask,wei2021finetuned,chung2022scaling,taori2023alpaca}. 

\begin{figure}[t]
    \centering
    \includegraphics[width=0.99\linewidth, trim=0 0 0 0]{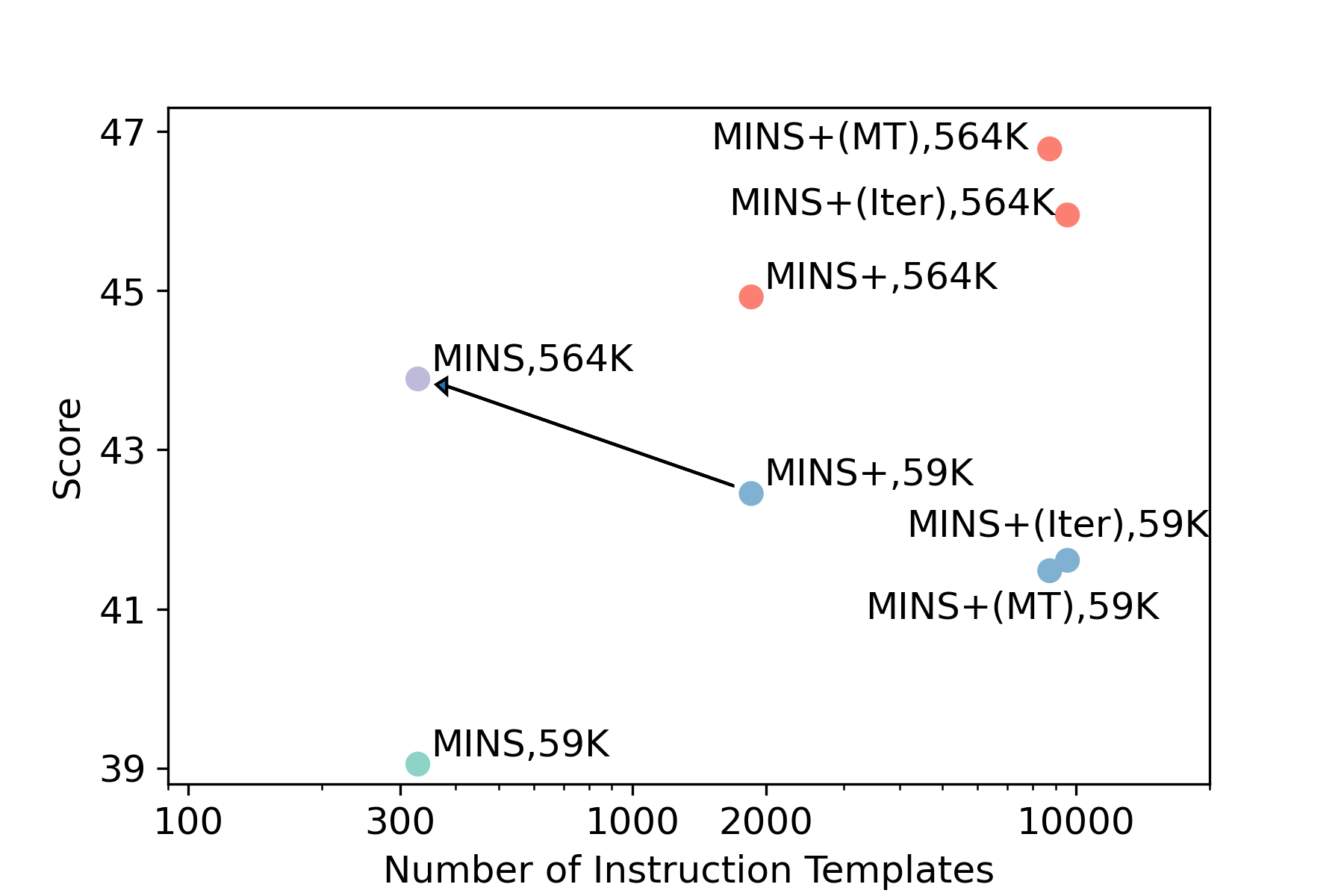}
    \caption{Zero-shot performance on~\textsc{MultiInstruct}~test set (9 tasks) by OFA tuned on each instruction-following dataset. 
    By expanding the instruction set several times using automatically generated instructions~(``MINS,59K'' to ``MINS+,59K''), the average score is close to that tuned using 10x more data (``MINS+,59K'' compared to ``MINS,564K'', highlighted by the arrow).}
    \label{fig:teaser}
\end{figure}

When deploying the instruction-tuned LLMs on unseen tasks, users from various backgrounds may write instructions with diverse wording styles to test the performance of the same task. In light of this, high-quality instruction-following data covering a wide range of wording styles are required in the alignment of large language models.
The conventional approach to obtain a large instruction-following dataset resembles data annotation, which recruits crowdsourced workers who engage in NLP research and possess rich knowledge about the target tasks to write instructions for each task/instance~\cite{sanh2021multitask,wang2022super}. 
\citet{zhou2023lima}~further eliminates redundant annotations, proving that IFT on only well-crafted 1,000 instructions can align as huge as 65B models with user preferences. 
While achieving a comparable level to the state-of-the-art baselines like GPT-4~\cite{openai2023gpt}, these LLMs instruction-tuned by hand-written data inevitably necessitate numerous iterations of thoughtful consideration and checking, which cost substantial human labor.
Although recent works have explored automatic instruction generation through in-context learning~\cite{wang2022self, honovich2022unnatural,taori2023alpaca}, the generated instruction-data pairs still require meticulous identification and filtering by experienced workers to yield a practical dataset. 

The same issue exists in multimodal tasks. Multimodal IFT (MIFT) has demonstrated remarkable efficacy in addressing visual question answering tasks. Some of the instruction tuning datasets such as \origdatasetname~\cite{xu2022multiinstruct}~contain more than 500K instances but pretty few instructions. 
To break through the bottleneck, we argue that~\emph{Enriching the training data with diverse task-specific instructions can enhance the effectiveness of MLLMs to a range of zero-shot task instructions}~and propose a completely automatic instruction-following dataset expansion framework~\methodname~for MIFT.
By harnessing an LLM and starting from straightforward \textit{meta-prompts}~\cite{suzgun2024metaprompting}, we expand the hand-written instruction templates to a size~\textbf{30 times larger}.
We fine-tune the two base models 
on the augmented dataset that incorporates the generated instruction templates. 
The results on two benchmarks that cover a broad range of task types manifest that~\methodname~can facilitate the IFT learning process, particularly in scenarios where the training data is limited. 
We find augmenting the dataset with about 1500 generated instructions can boost MLLM's performance by 2-3\%.
Moreover, it exhibits better adaptability to instructions in different language styles and nearly catches up with the result by expanding the dataset 10 times~(\Cref{fig:teaser}).
To put it in a nutshell, our contributions are as follows:
\begin{itemize}
    \item We present~\methodname, an automatic and labour-free dataset expansion framework for MIFT.
    \methodname~relies purely on rule-based filters to remove invalid generations and adopts a self-adaptive sampling strategy to balance instructions' consistency and diversity in the resulting dataset.
    Compared to previous peer works,~\methodname~almost requires no human labor and can be done by a single person.
    \item We expand a raw MIFT dataset~\textsc{MultiInstruct}~(\textsc{MIns})~to~\datasetname~(this name means it is an upgraded version of~\textsc{MultiInstruct}) using~\methodname. Fine-tuning two MLLMs on~\datasetname~yields promising results on different task types whose instruction templates cover a broad range of wording styles.
    \item We further conduct a comprehensive empirical study based on the obtained results and propose our insights from multiple aspects.
\end{itemize}


\section{Related Work}
\subsection{Instruction Finetuning on MLLMs}
Finetuning pretrained large language models on
human-written multi-task instruction-following data has shown unprecedented power for zero-shot task generalization as well as alignment with human intentions to accomplish real-world tasks~\cite{wei2021finetuned,wang2022self, wang2022super,ouyang2022training,longpre2023flan,liu2023visual}.
As large visual foundation models like Flamingo~\cite{alayrac2022flamingo}, BLIP~\cite{li2022blip,li2023blip} and SegmentAnything~\cite{kirillov2023segment} emerge, many visual-language tasks can be formulated in a sequence-to-sequence prediction manner and effectively solved by these models, and consequently, IFT has been successfully extended to the multimodal field and its efficacy has been verified~\cite{liu2023visual,zhang2023llama,dai2305instructblip,han2023imagebind,li2023mimic,han2023sas}.
Most of these works employ rule-guided human-written instructions through crowdsource workers, which are not cost-effective if the required instruction size is huge.

\subsection{Automatic Instruction Generation}
Since instruction curating is costly nowadays, researchers seek automatic data generation methods to cut down the expenses. Unnatural instructions~\cite{honovich2022unnatural} and Self-instruct~\cite{wang2022self} firstly exploit GPT-3 (text-davinci) and in-context learning to generate new I/O pairs and corresponding instructions.
This method has been applied to produce more advanced instruction-following models like Alpaca~\cite{taori2023alpaca}~and LLaMA-adapter~\cite{zhang2023llama}.
~\citet{li2023self}~train a model to generate instructions inversely for the input instances and iteratively selects high-quality data to train this instruction generation model the in next stage.
~\citet{zhou2023lima}~prove as few as 1,000 carefully curated instructions are sufficient to yield decent zero-shot performance on downstream tasks.
These works inevitably utilize a number of trained human annotators.
The other line of works~\cite{xu2023wizardlm,sun2023principle,liu2023makes} proposed to use a few conceived abstract principles or augmentation guidelines to refine existing instructions in order to obtain model-understandable instructions. However, these works concentrate on the augmentation of input text by increasing the difficulty or broadening the question scope instead of task description data, which may not be directly applied to augment task descriptions.
\section{Method}
\subsection{Multimodal Instruction Fine-tuning (MIFT)} 
MIFT uses instruction-following data in the training set to tune a base model $\mathcal{M}$ with parameter $\theta$, and expects the model to perform well on the unseen test tasks.
Formally, given the MIFT dataset $D=\{\mathcal{D}^{train}, \mathcal{D}^{test}\}$, where $\mathcal{D}^*=\{\mathcal{D}_1^*, \mathcal{D}_2^*,...,\mathcal{D}_{\vert \mathcal{D}^{*}\vert}^*\}$ is the dataset collection of $\vert \mathcal{D}_* \vert$ tasks in the individual split. 
$\mathcal{D}^{train} \cap \mathcal{D}^{test} = \oslash$ to satisfy the zero-shot setting.
Each task-specific dataset $\mathcal{D}_k=\{(\textbf{x}_i, \textbf{y}_i)\}_{i=1}^{N_k} \cup \{\mathcal{T}_j\}_{j=1}^{M_k}$ consists of input-output pairs $(\textbf{x}_i, \textbf{y}_i)$ and an instruction template sets $\{\mathcal{T}_j\}_{j=1}^{M_k}$, where $M_k \geq 1$ means there is usually more than one template for each task. 
Each instance has been~\textit{instantiated}~before being fed to the base model $\mathcal{M}$ using instruction template $\mathcal{T}_j$:
\begin{equation}
\mathbf{x}'_i=\{\mathcal{T}_j(\mathbf{x}_i)\}
\end{equation}
During tuning procedure, we maximize the likelihood of ground truth sequence of length $T$ in an auto-regressive fashion:
\begin{equation}
    \mathcal{L} = - \sum_{t=1}^{T}\log P_\theta(\mathbf{y}_{i,t}|\mathbf{x}'_i,\mathbf{y}_{i,<t})
\end{equation}
The model is then assessed on $\mathcal{D}_{test}$, where the score is calculated with the task-specific criterion $\mathcal{C}$ between predicted output $\hat{\mathbf{y}}_i^k$ and truth tokens $\mathbf{y}_i^k$ and averaged across the dataset $\mathcal{D}_k$.
\begin{equation}
    score_k = \frac{1}{N_k}\sum_i\mathcal{C}_k(\hat{\mathbf{y}_i}, \mathbf{y}_i) 
\end{equation}

\begin{figure*}
    \centering
    \includegraphics[width=0.92\textwidth, trim=0 1.2cm 0 0]{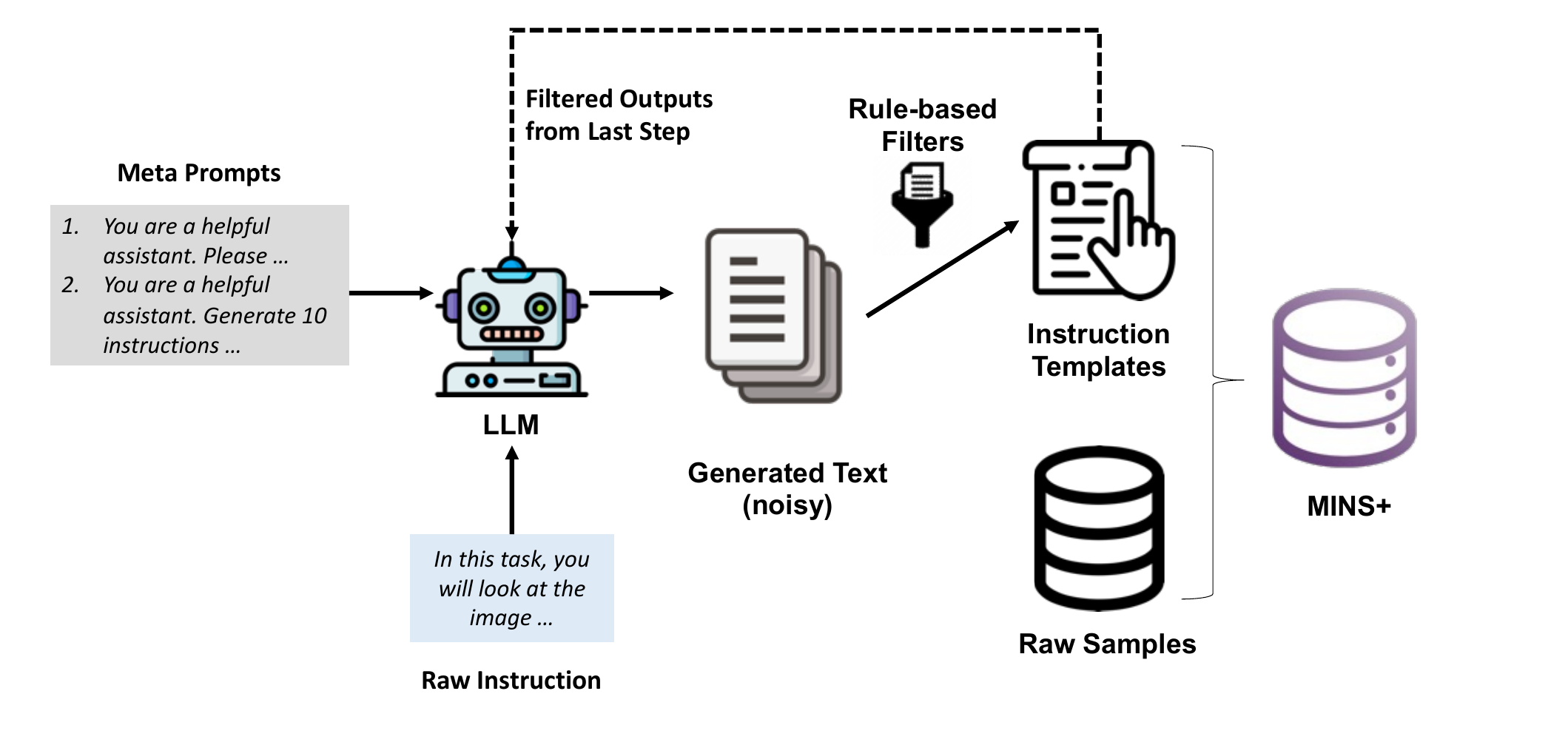}
    \caption{The overall framework of~\methodname. The self-loop only iterates once when we roll out instructions from the meta-prompt (e.g., ``generate 10 instruction about ...'').}
    \label{fig:framework}
\end{figure*}

\subsection{Generation Process}
The overall pipeline of our proposed framework is illustrated in \Cref{fig:framework}. 
At the beginning, we initialize the LLM with meta-prompts, which provide guiding instructions to steer the LLM in generating new instructions. 
The inputs to the LLMs comprise both the guiding instructions and original instructions, encompassing raw instructions from \textsc{MIns} (solid line) and filtered valid outputs from the preceding step (dashed line, as we consider only valid inputs likely to yield valid outputs). Upon gathering the outputs from the LLM, we apply a sequence of concise rule-based filters to eliminate invalid content and inconsistent instructions. Subsequently, we construct \textsc{MIns+} using the generated instruction templates and the raw instances. Finally, we fine-tune the base model on \textsc{MIns+} (not depicted in the figure).

\subsubsection{Meta-Prompts}
We explore two types of meta-prompts: \emph{direct prompting} and \emph{bootstrapping}.
Direct prompting involves providing a clear, human-conceived prompt that explicitly communicates the objective of generating new instructions from the given one to the LLM. For instance, an example prompt could be, "\textit{Rephrase the given instruction}."
In contrast, bootstrapping compels the LLM to generate a predetermined number (e.g., 10 in our experiments) of instructions that can guide the LLM in generating direct prompting instructions. 
We provide some input-output examples  in~\Cref{sec:guiding_gen}.

For both types of meta-prompts, we adhere to the common practice~\cite{touvron2023llama2} of prepending a system prompt such as "\textit{You are a helpful assistant.}" (\Cref{fig:framework}) to the generation prompt. The generation procedure can iterate and repeat numerous times, where the valid output from the last step serves as the input in the new iteration.
As the loop proceeds, the occurrence of duplicated outputs increases. Consequently, we halt the process after several iterations once we have amassed the desired number of instructions. Detailed prompt template formatting and response examples are available in the appendix.

In addition to iterative generation, inspired by~\citet{wang2022self}, we also experiment with sampling from different temperatures to further diversify the instruction set. In subsequent discussions, we refer to this variant as $\datasetname_{\text{MT}}$, while the iterative collection method is denoted as $\datasetname_{\text{Iter}}$.

\begin{algorithm}[ht]
\small
\SetKwFunction{ret}{return}
\KwIn{Input instruction template $I_{old}$ with placeholders. Large language model $\mathcal{M}$.}
\KwOut{New instruction template $I_{new}$}
\tcc{Find all unique placeholders}
$\mathcal{S}_{ph} \leftarrow FindPlaceHolders(I_{old})$ \\
\tcc{Mask placeholders with atomic substitutions}
$MaskList \leftarrow \{``\{A\}",``\{B\}",...\}$ \\
$bmdict \leftarrow \{\}$ \tcp{backmapping dict (masks to placeholders)} 
$ I \leftarrow I_{old}, i \leftarrow 0$ \\
\For{$ph \in \mathcal{S}_{ph}$}{
    $I \leftarrow Sub(I, ph, MaskList[i])$ \\
    $bmdict[MaskList[i]] = ph$ \\
    $i \leftarrow i+1$
}
\tcc{Generate with prompts}
$\hat{I} \leftarrow \mathcal{M}(P_{gen},I)$ \\
\tcc{Restore placeholders}
\For{$Mask, ph \in bmdict$}{
    $\hat{I} \leftarrow Sub(\hat{I},Mask,ph)$
}
$I_{new} \leftarrow \hat{I}$ \\
\Return $I_{new}$
\caption{Placeholder-Protected Generation (PPG)}\label{alg:ppg}
\end{algorithm}

\subsubsection{Handling Placeholders}
One challenge during generation is how to deal with instructions with \textit{placeholders}---blanks to be filled during instantiating time. 
Such placeholders can be simply expressed as format string variables in Python, but can not be properly handled by an LLM.
For example, one of the prompts for the Grounded Captioning task is ``\textit{Describe the content of} \rm{\{region\_split\_token.join(region)\}} \textit{in the picture.}'', where ``region\_split\_token'' and ``region'' are item names from an instance and can be initialized as the format string in Python when loading the instance.
If we directly ask the model to rewrite, the output could be ``\textit{Give a description for}~\{list of region\}'', since the LLM may treat the content within the brackets as regular words/phrases and rewrite them as well. 
This outcome incurs value error when built with examples as format strings in the code.

To alleviate this issue, we devise a Placeholder-Protected Generation (PPG) algorithm that combines the technique of masking and controlled generation, as illustrated in~\Cref{alg:ppg}. 
We substitute placeholders with a predefined list of ``atomic'' (i.e., hardly to be modified in generation) masks before feeding them into the language model, then we map them back to these placeholders in the generation results.
Specifically, we found single alphabet has promising atomic properties. Hence we choose these simple elements as protectors.
During generation time, we append the restriction command ``\textit{Keep the content within brackets (including `\{\}') unchanged}.'' to the generation instruction.
By combining these two add-ons, the model can produce valid augmentations in most cases. 

\subsubsection{Postprocessing}
We apply the following rule-based filters to the primary results:
1) Duplication examination and removal through perfect match.
2) For original instructions containing placeholders, we preserve those in an~\emph{unordered}~match with the original ones.
3) Length-based filters to overcome non-aligned responses since we found these responses are much longer than the normal ones (See~\Cref{sec:hallucination}~for examples). 
We find this hallucination removal filter is extremely effective when the instruction set is large (See our ablation study in~\Cref{sec:abl_study}).
Finally, we gather all valid instructions for the building of the augmented dataset.
The details of the result dataset can be found in~\Cref{sec:mins_detail}.

\subsection{Fine-tuning on~\datasetname}
\subsubsection{Source Dataset}
We curate~\datasetname~by sampling from the training dataset collection of~\textsc{MultiInstruct}~\cite{xu2022multiinstruct}, which consists of 62 tasks (53 training tasks and 9 evaluation tasks) based on 21 existing datasets. 
We use the same split of training and testing tasks as in~\textsc{MultiInstruct}.
Since it is unknown how many instances exactly sampled from each task, we respectively sample two versions in~\textsc{MIns+}---1K and 10K instances per task (59K and 564K in total), which also provides an opportunity to study our method's effectiveness under different dataset sizes. 
Unless specifically mentioned, each instance only appears once in the final dataset.

\subsubsection{Dataset Construction}
\label{sec:sample}
The next step is to construct the new with generated instruction templates.
A straightforward practice is to randomly pick one instruction template (with repetition) and one instance (without repetition) from the pools and combine them.
However, from the perspective of data augmentation, this model-agnostic method neglects the distribution of these instruction templates in the MLLM's textual encoder's hidden space.
Therefore, we develop an adaptive sampling procedure within each task's scope.
Concretely, we assign probability (a preset hyperparameter) $\epsilon$ and $1-\epsilon$ to the original and generated instruction set $\mathcal{I}_{orig}$ and $\mathcal{I}_{gen}$, respectively, where $\epsilon$ is a preset parameter and $\vert \mathcal{I}_* \vert$ is the cardinality of set $\mathcal{I}_*$.
For original instructions, we deem they contribute equally to the training so that the probability to pick each template from $\mathcal{I}_{orig}^k$ for task $k$ is $\frac{\epsilon^k}{\vert \mathcal{I}_{orig}^k \vert}$.
For generated instructions, we calculate for each of them a score $s$, and the probability is over a softmax normalization.
By default $\epsilon^k = \frac{\lvert \mathcal{I}_{orig}^k \rvert}{\lvert \mathcal{I}_{orig}^k \rvert + \lvert \mathcal{I}_{gen}^k \rvert}$ in our experiments to ensure a balanced mapping.
\begin{equation}
    P(I_{j}) =
    \begin{dcases*}
        \frac{\epsilon^k}{\vert \mathcal{I}^k_{orig} \vert} & if $I_{j} \in \mathcal{I}_{orig}^k$, \\[1ex]
        \frac{\exp(s_j)\times(1-\epsilon^k)}{\sum_{j'\in \vert I^k_{gen} \vert} \exp{(s_{j'})}} & if $I_{j} \in \mathcal{I}_{gen}^k$
    \end{dcases*}
\end{equation}

To compute $s_j$, a formulation should balance well between~\textit{consistency}~and~\textit{diversity}~\cite{sanh2021multitask}.~\textit{Consistency}~means that the generated instruction can convey precise task description as its origin.~\textit{Diversity}~denotes that the training procedure has a comprehensive coverage over the instruction wording space so that for various instruction templates the model can produce similar results.
Combining these two factors, we use the following expression to score each generated instruction:
\begin{equation}
    s_j = \mathbf{sim}(E_j,E_{orig}) - \frac{1}{N}\sum \mathbf{sim}(E_j, E_{j'})
\end{equation}
where $\mathbf{sim}$ is the similarity function, $E_*$ is the embedding of $I_*$ produced from the MLLM's text encoder. The first item quantifies the similarity between the generated instruction and its origin (consistency), while the second item measures the divergence (negative similarity, or diversity) between the instruction and $N$ randomly sampled counterparts $I_{j'}$ generated from the same $I_{orig}$.
The formulation of this sampling strategy resembles contrastive feature learning or selection~\cite{han2021improving,han2022mm}~and has shown its effectiveness in many application scenarios.



\section{Experiments}
\label{sec:exp}
\subsection{Experiment Setup}
\paragraph{Models}
we use Llama-2-13b-chat and GPT-4 for revised instruction generation.
Our target MLLMs for tuning include OFA~\cite{wang2022ofa}, Instruct-BLIP~\cite{dai2305instructblip}~and~LLaVA~\cite{liu2023visual,liu2023improved}. More details of these models can be found in~\Cref{appendix:model}.

\paragraph{Tasks and Metrics}
We select tasks from three representative MIFT benchmarks,~\textsc{MultiInstruct}~\cite{xu2022multiinstruct}~(9 tasks),~InstructBLIP~\cite{dai2305instructblip}~(3 tasks)~and~MMMU~\cite{yue2023mmmu}~for evaluation, as shown in~\Cref{tab:eval_tasks}. 
Among these tasks, VSR~\cite{liu2023visual}~and Visdial~\cite{das2017visual}~are included in both benchmarks (counted as~\textsc{MultiInstruct}~tasks).
We test for extant templates on~\textsc{MIns}~tasks,
which can be found in~\cite{xu2022multiinstruct}.
For IBLIP-Bench multiple-choice tasks (ScienceQA and IconQA), we reconstruct another template in~\textsc{MIns}~style for these tasks based on the templates from the IBLIP paper~\cite{dai2305instructblip}. 
The detailed templates can be found in~\Cref{sec:eval_templates}.

\paragraph{Baseline Method}
Models tuned on the primitive version of~\textsc{MIns}~serve as a basic baseline for comparison. 
Additionally, we follow~\cite{zhu2023promptbench}~to create adversarial prompts from 4 aspects for tuning, dubbed~\textit{Adv-Train}~as another baseline in the later context.

\begin{table*}
\resizebox{\linewidth}{!}{
    \begin{tabular}{c|l|l|*{11}{c}|l}
        \toprule[2pt]
        \multirow{2}{*}{Model} & \multirow{2}{*}{Size} & \multirow{2}{*}{Dataset} &VTE & TVQA & VisDial && CVQA & DC & GVQA & VE & VSR & NLVR && \multicolumn{1}{l}{\multirow{2}{*}{Avg.}} \\ 
        && ~ & \multicolumn{3}{c}{\cellcolor{Gray} \textbf{Rouge-L}} && 
        \multicolumn{6}{c}{\cellcolor{Gray} \textbf{Acc.}} && ~
        \\
        \midrule
         \multirow{8}{*}{OFA} & \multirow{4}{*}{59K}  & \textsc{MIns} & 29.44 & 20.75 & \underline{35.14}  && \underline{23.24} & 49.28 & 20.48 & 54.22 & 56.27 & 62.60 && 39.05 \\
         && \textsc{MIns+} & \textbf{35.33} & \textbf{21.52} & \textbf{35.55} && 22.70 & \textbf{54.94} & \textbf{20.63} & \underline{54.24} & 67.48 & \underline{69.64} && \textbf{42.45}~\inc{3.40} \\
         && $\textsc{MIns+}_\textrm{Iter}$ & 30.11 & \underline{20.83} & 34.32 && 23.16 & 51.94 & \underline{20.62} & 54.12 & \underline{67.59} & \textbf{71.77} && \underline{41.61}~\inc{2.56} \\
         && $\textsc{MIns+}_\textrm{MT}$ & \underline{31.50} & 20.74 & 33.30 && \textbf{23.28} & \underline{52.44} & 20.38 & \textbf{54.48} & \textbf{67.83} & 69.36 && 41.48~\inc{2.43} \\
         \cline{2-15}
         & \multirow{4}{*}{564K} & \textsc{MIns} & 41.33 & 24.49 & \textbf{36.18} && 26.84 & 47.68 & 49.25 & 50.74 & 63.96 & 54.52 && 43.89 \\
         && \textsc{MIns+} & 45.43 & \textbf{24.62} & 34.98 && \underline{29.16} & 51.06 & 49.64 & \underline{52.76} & \textbf{66.74} & \underline{54.70} && 44.92~\inc{1.03} \\
         && $\textsc{MIns+}_\textrm{Iter}$ & \underline{45.63} & 23.36 & 34.30 && \textbf{30.02} & \textbf{53.84} & \underline{53.73} & 52.18 & 66.11 & 54.34 && \underline{45.95}~\inc{2.06} \\
         && $\textsc{MIns+}_\textrm{MT}$ & \textbf{51.71} & \underline{24.55} & \underline{35.08} && 27.52 & \underline{52.24} & \textbf{56.12} & \textbf{53.06} & \underline{66.72} & \textbf{55.74} && \textbf{46.78}~\inc{2.89}\\
         \midrule[2pt]

         \multirow{8}{*}{IBLIP} & \multirow{4}{*}{59K} &  \textsc{MIns} & - & 17.14 & - && 27.46 & 17.44 & 24.00 & 51.76 & 53.60 & 54.12 && 35.07 \\
         && \textsc{MIns+} & - & 17.84 & - && 27.54 & \underline{22.46} & \textbf{25.30} & 53.30 & \textbf{56.08} & 56.28  && \underline{36.97}~\inc{1.90} \\
         && $\textsc{MIns+}_\textrm{Iter}$ & - & \underline{18.01} & - && \textbf{28.10} & 21.10 & 24.14 & \underline{53.64} & 54.74 & \underline{56.48} && 36.60~\inc{1.53} \\
         && $\textsc{MIns+}_\textrm{MT}$ & - & \textbf{18.29} & - && \underline{27.62} & \textbf{22.72} & \underline{25.36} & \textbf{54.84} & \underline{55.66} & \textbf{56.80} && \textbf{37.33}~\inc{2.26} \\
         \cline{2-15}
         
         & \multirow{4}{*}{564K} & \textsc{MIns} & - & 15.12 & - && 27.66 & 29.28 & 24.20 & 50.44 & 52.84 & 56.26 && 36.87 \\
         && \textsc{MIns+} & - & \textbf{17.19} & - && \textbf{29.24} & \underline{34.82} & 25.48 & \underline{54.56} & \textbf{57.16} & \textbf{57.30} && \textbf{39.39}~\inc{2.52} \\
         && $\textsc{MIns+}_\textrm{Iter}$ & - & \underline{16.67} & - && \underline{27.90} & 33.30 & \textbf{26.36} & 53.90 & 56.06 & 56.81 && 38.71~\inc{1.84} \\
         && $\textsc{MIns+}_\textrm{MT}$ & - & 16.09 & - && 27.88 & \textbf{36.48} & \underline{25.86} & \textbf{55.34} & \underline{56.52} & \underline{57.22} && \underline{39.34}~\inc{2.47} \\
         \bottomrule[2pt]
    \end{tabular}
    }
    \caption{Results on~\textsc{MultiInstruct}~test bench (9 tasks). The reported digits for each task are averaged over 5 hand-written instructions. The colored digits surrounded by brackets are relative gain from finetuned~\textsc{MIns}~baseline in the corresponding sizes. The \textbf{highest} and \underline{second highest} scores are marked~\textbf{in bold}~and~\underline{with underline}. Results of VTE and VisDial on IBLIP are omitted due to task and data overlap with pretraining/tuning (see~\Cref{tab:eval_tasks}). 
    }
    \label{tab:res_mins}
\end{table*}
\subsection{Main Results}
\paragraph{Results acorss MLLMs}
We report the results of~\textsc{MIns}~tasks on OFA, InstructBLIP and LLaVA in~\Cref{tab:res_mins,tab:res_iblip,tab:res_llava}. 
There are several key findings observed from these tables:

First, finetuning both models on~\textsc{MIns}~and \datasetname~acquires significant gain on the held-out tasks over base models, demonstrating that MIFT is an effective learning paradigm for task generalization. 

Second, \datasetname~consistently improves~\textsc{MIns} on nearly all benchmarks and across all base models, which shows that tuning on datasets composed of both raw and composed instructions casts a positive effect on the completion of downstream tasks.
A surprising discovery is that the result of~\datasetname~(59K) is competitive with (42.45 vs. 43.89, 1.44 behind OFA-Large) and even surpasses (37.33 vs. 36.87, 0.46 points ahead of IBLIP) that of~\textsc{MIns}~(564K), implying that tuning with expanded instruction set can yield similar or better overcome as dataset expansion (approximate 10$\times$ times).
\begin{table}
\resizebox{\linewidth}{!}{
    \begin{tabular}{c|l|*{4}{c} l}
        \toprule
        Model & Method & SciQA & IconQA & VizWiz  &&  \multicolumn{1}{l}{Avg.} \\ 
        \midrule
        0 & OFA-Large & 3.90 & 16.92 & 7.76 && 12.42 \\
        \midrule
         \multirow{4}{*}{OFA} &  \hspace{4pt}+\textsc{MIns} & 26.18 & \underline{38.70} & 18.67 && 27.85 \\
         ~ & \hspace{4pt}+\textsc{MIns+} & \textbf{27.32} & 38.25 & \textbf{19.88}  && \textbf{28.48}~\inc{0.63} \\
         ~ & \hspace{4pt}+$\textsc{MIns+}_\textrm{Iter}$ & \underline{26.69} & 38.45 & 18.94 && 28.03~\inc{0.18} \\
         ~ & \hspace{4pt}+$\textsc{MIns+}_\textrm{MT}$ & 26.41 & \textbf{39.40} & \underline{19.39}  && \underline{28.40}~\inc{0.55} \\
        \midrule
        0 & IBLIP & 38.29 & 37.71 & 12.19 && 29.40 \\
         \midrule
         \multirow{4}{*}{IBLIP} &  \hspace{4pt}+\textsc{MIns} & 49.03 & 39.85 & 24.73 && 37.87  \\
         ~ & \hspace{4pt}+\textsc{MIns+} & \textbf{49.75} & \textbf{40.22} & \underline{27.87} && \textbf{39.28}~\inc{1.41} \\
         ~ & \hspace{4pt}+$\textsc{MIns+}_\textrm{Iter}$ & \underline{49.26} & 39.91 & \textbf{27.96} && \underline{39.04}~\inc{1.17} \\
         ~ & \hspace{4pt}+$\textsc{MIns+}_\textrm{MT}$ & 49.35 & \underline{40.17} & 26.12 && 38.55~\inc{0.68} \\
         \bottomrule
    \end{tabular}
    }
    \caption{Results on three tasks from InstructBLIP evaluation bench tuned on OFA/IBLIP using 564K training data. The digits are averaged over IBLIP and~\textsc{MIns}~instruction templates.}
    \label{tab:res_iblip}
\end{table}

\paragraph{Comparison with other baselines}
We continue to verify~\methodname's effectiveness by comparing with other expansion methods as well as LLM assistants, as shown in~\Cref{tab:res_bl_comp}. 
\methodname~significantly outperforms the adversarial training approach (Adv-Train) across all three models, which manifests that~\methodname~has better capability of improving MLLM's robustness to versatile instruction styles. 
Besides, it is noteworthy that the evaluation results remain very close after switching the core LLM in~\Cref{fig:framework}~to more advanced GPT-4. 
This suggests that the proposed approach is convenient and cost-effective, as a random picked open-sourced model can ahieve gains equivalent to thos of the state-of-the-art proprietary counterpart.

\begin{table}
\centering
\resizebox{0.75\linewidth}{!}{
    \begin{tabular}{l|*{3}{l}}
        \toprule
         Method & \origdatasetname & IBLIP & MMMU \\ 
        \midrule
            \textsc{MIns} & 38.64 & 43.88 & 37.33 \\
            \textsc{MIns+} & \textbf{40.25} & \textbf{46.87} & \underline{38.59} \\
            $\textsc{MIns+}_\textrm{Iter}$ & 39.85 & \underline{46.53} & \textbf{38.71}  \\
            $\textsc{MIns+}_\textrm{MT}$ & \underline{40.02} & 46.07 & 38.42 \\
         \bottomrule
    \end{tabular}
    }
    \caption{Results on three benchmarks on LLaVA using 59K tuning set.}
    \label{tab:res_llava}
\end{table}
\begin{table}
\centering
\resizebox{0.98\linewidth}{!}{
    \begin{tabular}{l|*{3}{c}}
        \toprule
         Method & OFA & IBLIP & LLaVA \\ 
        \midrule
         \textsc{MIns} & 39.05 & 35.07 & 38.64 \\
         Adv-Train & 39.71 & 35.58 & 37.27 \\
         \textsc{MIns+}~(Llama-2-13b-chat) & \textbf{42.45} & \underline{36.97} & \textbf{40.25} \\
         \textsc{MIns+}~(GPT-4) & \underline{42.33} & \textbf{37.12}  & \underline{40.01} \\
         \bottomrule
    \end{tabular}
    }
    \caption{Performance of three MLLMs on~\textsc{MIns}~benchmark~tuned on~\textsc{MIns}~training set (59K) and its several expanded versions.}
    \label{tab:res_bl_comp}
\end{table}

\subsection{Scaling Laws among Results}
\label{sec:scaling}
We also notice the scaling laws as mentioned in~\citet{hoffmann2022training}~in terms of the dataset (including instruction set and number of samples) and model size.
\paragraph{Dataset}
As shown in~\Cref{tab:res_mins}, the performance can be enhanced by expanding the size of either instruction set (\textsc{MIns}~to~\datasetname) or instance set (59K to 564K) for tuning.
Under the same instance set size, using a larger instruction set can boost the overall performance (\textsc{MIns}~to~\datasetname). 
Nevertheless, the marginal gain gradually decreases and can even be reverted to negative. For example, the marginal gain on OFA 59K, from~\textsc{MIns}~to~\datasetname~is 3.40, but is 
$-0.84$ 
and 
$-0.97$ 
from~\datasetname~to~\datasetitername~and~\datasetmtname, respectively.
In contrast, under the same instruction set~(\datasetname~and~$\datasetname_\textrm{Iter/MT}$), MLLM's scores significantly elevate in a larger instance set. 
This outcome perhaps stems from more precise human-annotated datasets compared to automatically generated instructions that roughly match the original meaning, implying that data are more essential and instructions are auxiliary ingredients for an IFT training set.

\paragraph{Model}
The scaling factor similarly exists in terms of MLLM size. 
Leveraging the 59K instance set, the 870M OFA earns even higher improvement than 7B+ IBLIP on both raw~\textsc{MIns}~and instruction-augmented~(\datasetname~series) tuning set, indicating that a tiny dataset composed of diverse instructions is sufficient for smaller models to achieve decent performance, while larger models (IBLIP) requires a larger dataset (564K) to fulfill similar improvement.

\begin{table}[ht!]
    \centering
    \resizebox{\linewidth}{!}{
        \begin{tabular}{l|c c c}
        \toprule
        Setting & \datasetname & \datasetitername & \datasetmtname \\
        \hline
        OFA-Large (59K) & 42.45 & 41.61 & 41.48  \\
        \hspace{10pt}+Ordered Match & 41.87~\dec{0.58} & 41.24~\dec{0.37} & 41.05~\dec{0.43} \\
        \hspace{10pt}-Length Filter & 42.38~\dec{0.07} & 40.85~\dec{0.76} & 40.67~\dec{0.81} \\
        \hline
        OFA-Large (564K) & 44.92 & 45.95 & 46.78 \\
        \hspace{10pt}+Ordered Match & 44.21~\dec{0.71} & 45.16~\dec{0.79} & 45.81~\dec{0.97} \\
        \hspace{10pt}-Length Filter & 44.77~\dec{0.15} & 44.99~\dec{0.96} & 45.61~\dec{1.17} \\
        \bottomrule
        \end{tabular}
    }   
    \caption{Ablative experiments on~\textsc{MIns} test set. The digits are averaged on 9 tasks.}
    \label{tab:abl}
\end{table}

\subsection{Ablation Study}
\label{sec:abl_study}
We consider the following ablative settings regarding the post-processing filters: 1) For placeholder examination, we substitute the~\textit{unordered}~matching with ordered matching. That is, if placeholders of the raw instruction are in the order ``$..\{A\}..\{B\}..\{C\}..$'', then only the outputs organized in the same order are preserved, and those which look like ``$..\{B\}..\{A\}..\{C\}$'' and ``$..\{C\}..\{A\}..\{B\}..$'' will be eliminated. 
2) We remove the length filter to allow hallucinated output to participate in the construction of~\datasetname. 
As illustrated in~\Cref{tab:abl}, forcing ordered matching or discarding the length filter downgrades the overall performance, which implies that shuffling placeholders is an effective augmentation for the model, and the length filter removes harmful hallucinated outputs. 
Moreover, we note that the extent of negative impact is dependent on the augmented instruction set size. 
The drop is more severe when enforcing ordered match in~\datasetname, but   on~\datasetitername~and~\datasetmtname~is larger when disabling the hallucination filter.

\section{Further Analysis}
In this section, we first seek some task attributes as potential factors to impact the performance of each task on~\datasetname.
Then we explore how the sampling strategy in~\Cref{sec:sample}~can affect the overall performance on evaluation benchmarks.

\subsection{Potential Impact Factors to~\datasetname}
\label{sec:factor_explore}
\begin{figure}
    \centering
    \includegraphics[width=0.98\linewidth]{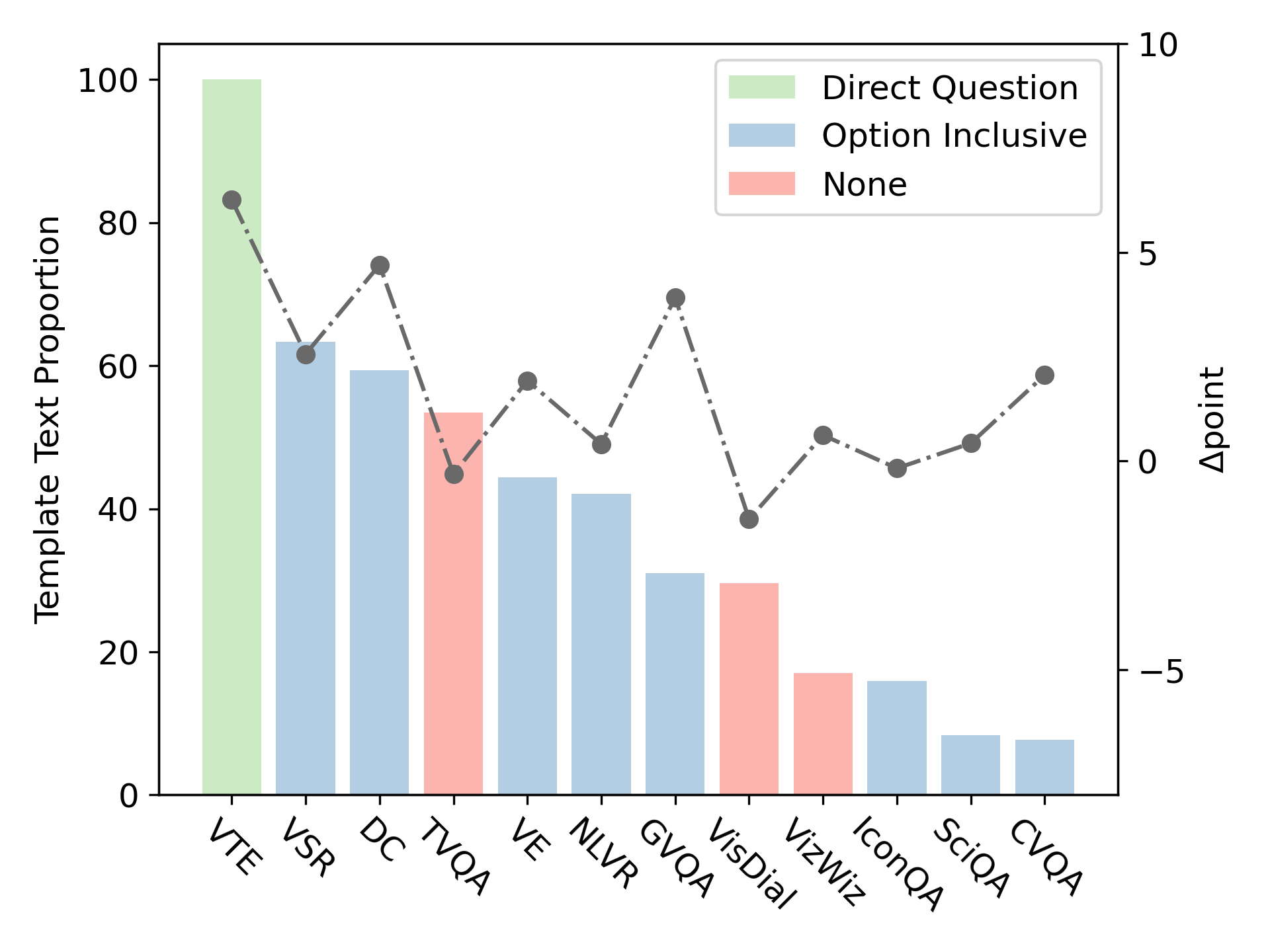}
    \caption{Average performance gain of three versions of~\datasetname~on OFA-Large vs. instance text proportion by task ($\rho=0.639$).}
    \label{fig:perf_task_attr}
\end{figure}

As mentioned above, \methodname~can improve the performance on the benchmark level.
Still, we are curious about what dominates and how much each individual task can benefit from instruction augmentation.
To probe the potential factors at the task level, we define the following task particulars:
\begin{itemize}
    \item direct\_question (bool): if the instruction template directly tells the model what it should respond to.
    For example, a template of image captioning ``Describe what you see in the image'' proposes the entire question to the task. Usually, this type of template does not contain any \textit{instance-specific} placeholders.
    \item option\_inclusive (bool): if the template provides a placeholder for options. This mainly appears in multi-choice question-answering tasks.
    \item template\_text\_proportion (float in $[0,1]$): the dataset-wise average proportion of template text length (excluding placeholders) in the entire \textit{instantiated}~instruction. 
\end{itemize}
We display the average gain for each task on OFA (59K) together with the individual attributes in~\Cref{fig:perf_task_attr}.
The Pearson Correlation Coefficient $\rho$ between instance text proportion and the average performance gain is 0.639, which uncovers a strong correlation between the relative length of instance text to the extra gain by~\methodname.
Moreover, tasks with shorter instance text fall behind those with longer instance text.
A probably reasonable explanation for this is that MLLMs attend excessively to the task-irrelevant templates, which could perturb the gradient to an inappropriate direction and scale~\cite{von2023transformers}. 

When the problem to solve has been entirely enclosed in the template like VTE, the task can benefit more from~\methodname, since for such tasks in the training set~\methodname~is imposed directly on the question rather than modifier text.
Compared to those whose instructions contain options after instantiation, tasks without visible options (TextVQA and VisDial) are susceptible to deterioration on~\datasetname, possibly because instruction augmentation can help consolidate the~\textit{copy-from-input}~ability by modifying the surrounding context, while pure generative tasks do not rely on this ability to predict answers.

\subsection{Sampling Strategy}
\begin{figure}
    \centering
    \includegraphics[width=\linewidth]{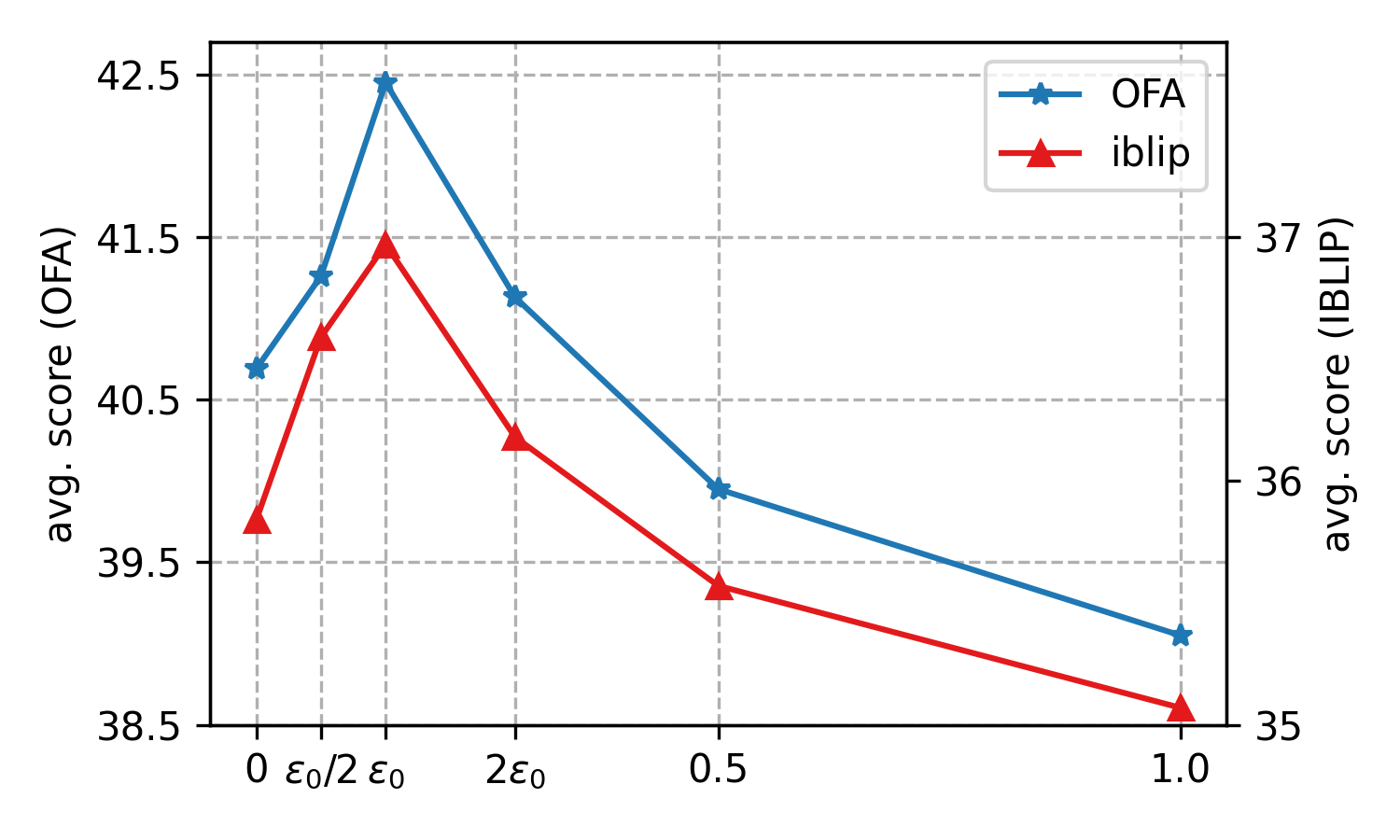}
    \caption{Results on~\datasetname~(59K) of different $\epsilon$ values for both models. $\epsilon=1.0$~is the result on~\textsc{MIns}.}
    \label{fig:perf_epsilon}
\end{figure}

We investigate the impact of the sampling strategy proposed in~\Cref{sec:sample}, mainly study on the hyperparameter $\epsilon$. 
There are 4 different $\epsilon$ values evaluated: $\{0,\epsilon_0/2,2\epsilon_0, 0.5\}$, where $\epsilon_0$ is the default task-specific value in our implementations.
Evaluations are conducted on~\datasetname~since the instruction set is sufficiently small to guarantee that $\epsilon_0/2$ and $2\epsilon_0$ deviate from $\epsilon_0$ distantly).
According to~\Cref{fig:perf_epsilon}, when sampling with a larger or smaller $\epsilon_0$, the average score on evaluated tasks decreases apparently.
The performance becomes even worse when using a fixed large value of $\epsilon$ for all tasks (0.5). 
These records reflect the importance of the balance between diversity and consistency in sampling, and our devised simple strategy reaches this goal.

\section{Conclusion}
In this paper, we propose~\methodname, a completely automatic instruction augmentation framework for large instruction-following datasets with very scarce hand-written instructions.
The framework purely relies on an arbitrary LLM and simple meta-prompts. 
We specially devise the generation process and rule-based filters based on the characteristics of raw instructions and model outputs.
Experimental results on two MLLMs across a broad range of tasks demonstrate our proposed method can significantly enhance the overall performance compared to baselines using non-augmented datasets, which also shows the improved robustness of human-written instructions in different wording styles.

\section*{Limitations}
In~\methodname, we apply a series of rule-based filters to remove invalid outputs that could do evil to the~dataset construction and tuning process, as well as a heuristic sampling strategy to balance the distribution of generated instructions in different versions of~\datasetname.
However,~\methodname~works on augmentation with less focus on other dataset surgery techniques such as data selection~\cite{xia2024less}~and dataset pruning~\cite{yang2022dataset}, which we believe may further improve the current performance by combining them with~\methodname. We hold this thought as our future research direction.

\section*{Ethical Considerations}
We clarify ethical considerations in two aspects. 
First,~\methodname~is a fully automatic augmentation framework and no crowdworker is hired. 
Second, regarding the generated process from LLM (LLaMA-2-13b-Chat and GPT-4), since they have experienced sufficient safety-tuning, and our instructions do not include deliberate attack to elicit harmful output, we believe there is extremely low risk in generating harmful information. 
We have thoroughly checked the augmented instruction set---apart from possible hallucinations, most of the output follows the guiding instruction and just modifies the extant instructions in~\textsc{MIns}.
Moreover, we provide reference inputs and outputs in the appendix to make the generation process transparent and public, and commit to release the instruction set for public inspection.

\bibliography{anthology,custom}
\bibliographystyle{acl_natbib}

\appendix
\section{Baseline Models}
\label{appendix:model}
In this section, we elaborate on the details of the models evaluated in~\Cref{sec:exp}.
\paragraph{OFA}~\cite{wang2022ofa} is a task-agnostic sequence-to-sequence learning framework, which supports both unimodal and multimodal tasks. It is pretrained on 20M publicly available image-text pairs but can show great transferability to unseen tasks and domains. In our experiments, following~\cite{xu2022multiinstruct}, we finetune OFA-large which contains 870M parameters during evaluations.

\paragraph{InstructBLIP} \cite{dai2305instructblip} is one of the earliest trial on MIFT. It is built on pretrained BLIP-2~\cite{li2023blip}~and Vicuna~\cite{vicuna2023}, a finetuned version of LLaMA on user-shared conversations.
InstructBLIP is tuned on a sampled subset of 15M instances covering 7 major tasks and 13 datasets. 
During the fine-tuning stage, only the fully connected layer which projects the instruction-aware image features into Vicuna's input space, is updated, while both the image-encoder and Query Transformer (Q-Former) in BLIP-2 and the upside LLM (Vicuna-7B) are frozen.

\paragraph{LLaVA}
\cite{liu2023visual,liu2023improved}
LLaVA represents a novel end-to-end trained large multimodal model. 
It connects vision encoder to a pretrained LLM Vicuna for general-purpose visual and language understanding.
There are total two training phases to produce LLaVA: 
1) Feature alignment (pretraining): this stage aligns the visual features output from the vision encoder to the language's hidden space. The pre-training is based on 558K subset of LAION-CC-SBU dataset~\cite{schuhmann2021laion}~with both LLM and visual encoder frozen.
2) End-to-end tuning: GPT-4 is employed to provide synthetic multimodal dialogue data, which is then tuned along with science QA benchmark to further enhance the instruction-following ability of LLaVA. 
It achieves impressive chat capabilities reminiscent of the multimodal GPT-4 and sets a new state-of-the-art accuracy on Science QA.

Due to limited computation resources, we use LoRA~\cite{hu2021lora} to finetune LLaVA-vicuna-7B-v1.6 on our generated instruction dataset.

\section{More Experimental Results}
We also test IBLIP (594K~\textsc{MIns}~versions) on the Massive Multi-discipline Multimodal Understanding and reasoning (MMMU) benchmark~\cite{yue2023mmmu}, as shown in~\Cref{tab:res_mmmu}.
We can see that IBLIP tuned on augmented~\textsc{MIns}~surpassses non-augmented baselines on both validation and test sets.
Notably, MMMU uses similar template style as IBLIP-bench for evaluation~\cite{yue2023mmmu}, but the gain on MMMU is more significant than IBLIP-bench~(\Cref{tab:res_iblip}), which provides a supplementary evidence that pure template augmentation has the chance to benefit tasks with very concise templates.  

\begin{table*}[!t]
\centering
\small
\resizebox{\linewidth}{!}{
    \begin{tabular}{@{}lcccccccc@{}}
    \toprule
    \textbf{} & \textbf{\begin{tabular}[c]{@{}c@{}}Validation \\  Overall\end{tabular}} & \textbf{\begin{tabular}[c]{@{}c@{}}Test \\  Overall\end{tabular} } & \textbf{\begin{tabular}[c]{@{}c@{}}Art \&\\  Design\end{tabular}} & \textbf{Business} & \textbf{Science} & \textbf{\begin{tabular}[c]{@{}c@{}}Health \& \\ Medicine\end{tabular}} & \textbf{\begin{tabular}[c]{@{}c@{}}Human. \&\\  Social Sci.\end{tabular}} & \textbf{\begin{tabular}[c]{@{}c@{}}Tech \&\\  Eng.\end{tabular}} \\
     & (900) & (10,500) & (1,163) & (1,428) & (2,426) & (1,752) & (947) & (2,784) \\ 
    \midrule
    OpenFlamingo2-9B \cite{awadalla2023openflamingo} & 28.7 & 26.3 & 31.7 & 23.5 & 26.3 & 26.3 & 27.9 & 25.1 \\
    Kosmos2 \cite{peng2023kosmos} & 24.4 & 26.6 & 28.8 & 23.7 & 26.6 & 27.2 & 26.3 & 26.8 \\
    Fuyu-8B~\cite{fuyu2023} & 27.9 & 27.4 & 29.9 & 27.0 & 25.6 & 27.0 & 32.5 & 26.4 \\
    MiniGPT4-Vicuna-13B~\cite{zhu2023minigpt} & 26.8 & 27.6 & 30.2 & 27.0 & 26.2 & 26.9 & 30.9 & 27.2 \\
    LLaMA-Adapter2-7B~\cite{zhang2023llama} & 29.8 & 27.7 & 35.2 & 25.4 & 25.6 & 30.0 & 29.1 & 25.7 \\
    Otter~\cite{li2023otter} & 32.2 & 29.1 & 37.4 & 24.0 & 24.1 & 29.6 & 35.9 & {\underline 30.2} \\
    CogVLM~\cite{wang2023cogvlm} & 32.1 & 30.1 & 38.0 & 25.6 & 25.1 & 31.2 & 41.5 & 28.9 \\
    InstructBLIP-T5-XL~\cite{dai2305instructblip} & 32.9 & 30.6 & 43.3 & 25.2 & 25.2 & 29.3 & 45.8 & 28.6 \\
    BLIP-2 FLAN-T5-XL \cite{li2023blip} & 34.4 & 31.0 & 43.0 & 25.6 & 25.1 & 31.8 & 48.0 & 27.8 \\
    Qwen-VL-7B~\cite{bai2023qwen} & \underline{35.9} & 32.9 & 47.7 & \underline{29.8} & 25.6 & 33.6 & 45.3 & \underline{30.2} \\
    LLaVA-1.5-13B~\cite{liu2023visual} & \textbf{36.4} & 33.6 & \textbf{49.8} & 28.2 & 25.9 & \textbf{34.9} & \textbf{54.7} & 28.3 \\
    InstructBLIP-T5-XXL~\cite{dai2305instructblip} & 35.7 & \underline{33.8} & 48.5 & \textbf{30.6} & \textbf{27.6} & 33.6 & 49.8 & 29.4 \\
    BLIP-2 FLAN-T5-XXL~\cite{li2023blip} & 35.4 & \textbf{34.0} & \underline{49.2} & 28.6 & \underline{27.3} & \underline{33.7} & \underline{51.5} & \textbf{30.4} \\ \midrule

    InstructBLIP-Vicuna7B~\cite{dai2305instructblip} & 27.6 & 24.8 & 29.0 & \textbf{25.9} & 22.2 & 25.0 & 25.4 & 24.6 \\ 
    \midrule
    \quad +\textsc{MIns} & 27.4 & 25.7 & 29.1 & 24.6 & 22.8 & 25.5 & \underline{31.6} & 25.3 \\
    \quad +\datasetname & 28.2 & \underline{26.8} & \textbf{34.3} & 23.9 & \underline{23.4} & 25.8 & \textbf{32.9} & \textbf{26.7}\\
    \quad +\datasetitername & \textbf{29.1} & \textbf{26.9} & \underline{33.5} & \underline{25.3} & 22.8 & \textbf{26.2} & 31.2 & 26.1 \\
    \quad +\datasetmtname & \underline{28.5} & 26.5 & 33.1 & 23.2 & \textbf{23.5} & \underline{26.1} & 30.6 & \underline{26.5} \\
    \bottomrule
    \end{tabular}%
}
\caption{Results of different models on the MMMU \textbf{validation} and \textbf{test set}. The model with best performance in each category is \textbf{in-bold}, and the second best is {\underline{underlined}}. *: results provided by the authors. Baseline results are excerpted from~\citet{yue2023mmmu}.}
\label{tab:res_mmmu}
\end{table*}

\section{Generation Details}
\subsection{Guiding Instruction Generation}
\label{sec:guiding_gen}
We first display the details of the guiding instruction generation process, during which we query our LLM with a simple meta-prompt, as shown in~\Cref{tab:meta_gen}. 
We omit the irrelevant output in front of and behind the 10 itemized guiding instructions.

\subsection{New Instruction Generation}
To leave an intuitive impression on the new instruction generation process. The generated instructions (``Gen'') are produced by feeding the corresponding guiding instruction from~\Cref{tab:meta_gen}~along with the raw instruction (``Raw'') into the LLM. Formally, we use the simple template in~\Cref{tab:gen_form}:

where ``Guiding Instruction'' is 1 of the 10 instructions that we generate from the first step, ``Constrain Instruction'' only appears when the input instructions have placeholders (it is masked through~\Cref{alg:ppg}) to enforce the model preserve highlighted masks, ``[TEXT]'' is a special token that works as the delimiter to help the model recognize the boundary of input text, ``Raw Instruction'' is the raw instruction to be modified.

\subsection{The Intuition behind Length Filter}
\label{sec:hallucination}
In the generation process, we found there are many cases that which LLM produces hallucinated output. For example, when asking the model to rewrite the following instruction:

\vspace{5pt}
\centerline{\textit{Generate some text to describe the image.}} 
\vspace{5pt}

\noindent Instead of generating a plausible paraphrasing, the output could be like this:
\vspace{5pt}

\parbox{0.9\linewidth}{
\textit{\hspace{11pt}The photograph depicts a majestic landscape with a serene lake in the foreground, surrounded by towering trees with vibrant foliage. The water's edge is adorned with smooth rocks and lush vegetation, creating a picturesque setting. In the distance, the sky is painted with hues of blue and white, adding a touch of elegance to the scene. The overall ambiance is one of tranquility and natural beauty, inviting the viewer to pause and appreciate the splendor of the outdoors.
}}

\vspace{8pt}

In this response text, the model describes an image instead response to our inquiry, and we do not present any image here. This is abnormal behaviour called \textbf{hallucination}~\cite{ji2023survey}. The text here is possibly seen by the model in advance during the pretraining or tuning stage. 
We particularly note that such hallucinated output is always much longer than the original instructions (78 and 7 in this example). Hence we can rule out those extremely long generated texts to eliminate hallucinations.

\begin{table*}[ht]
    \centering
    \resizebox{\textwidth}{!}{
        \begin{tabular}{l p{13cm}}
        \toprule
        \textbf{[System Prompt]} & You are a helpful assistant. \\[10pt]
        \textbf{[User]} & Generate 10 instructions about how to rephrase short text. Your response should be cononical and formatted as enumerations. \\[0.8cm]

        \textbf{[System]} & 
            \begin{tabular}[t]{@{}p{13cm}@{}}
                (Irrelevant output) \\\\
                1. \textbf{Use synonyms}: Replace specific words or phrases with their synonyms to convey the same
                meaning for the input. \\\\
                2. \textbf{Rearrange phrases for the input sentence}: Reorganize sentences or phrases to improve 
                readability or flow without changing the overall meaning. \\\\
                3. \textbf{Simplify language}: Use simpler vocabulary and sentence structures to make the text more 
                accessible and easier to understand. \\\\
                4. \textbf{Remove redundancies}: Eliminate unnecessary words or phrases that do not add to the 
                meaning of the text. \\\\
                5. \textbf{Use active voice}: Convert passive sentences to active ones to make the text 
                more engaging and easier to read. \\\\
                6. \textbf{Break up long sentences}: Divide long sentences into  shorter ones to improve readability and flow. \\\\
                7. \textbf{Use bullet points or numbered lists}: Organize information using bullet points or numbered lists to make it easier to read and understand. \\\\
                8. \textbf{Replace jargon with common terms}: Use plain language instead of technical jargon or industry-specific terms to make the text more accessible to a wider audience. \\\\
                9. \textbf{Use inclusive language}: Avoid using language that stereotypes or excludes certain groups of people. Instead, use language that is respectful and inclusive of all individuals. \\\\
                10. \textbf{Check for consistency}: Ensure that the text is consistent in terms of tone, style, and format throughout. \\\\
                (Irrelevant output)
            \end{tabular} \\
        \bottomrule
        \end{tabular}
    }
    \caption{10 rewriting instructions generated from LLaMA-Chat-13B. We highlight the title of each instruction in bold. Extra vertical space is used to clearly split each point.}
    \label{tab:meta_gen}
\end{table*}

\begin{table}[ht]
    \centering
    \begin{tabular}{p{0.95\linewidth}}
    \toprule
        Please follow the description of the given instruction to modify the input text behind the token [TEXT].
        \newline
        \\[6pt]
        \{Guiding Instruction\} \{Constrain Instruction\}
        \\[6pt]
        [TEXT]: \{Raw Instruction\}  \\
    \bottomrule
    \end{tabular}
    \caption{New instruction generation chat formulation.}
    \label{tab:gen_form}
\end{table}

\begin{table*}[t]
    \centering
    \resizebox{\textwidth}{!}{
        \begin{tabular}{c|c|c|m{0.6\textwidth}}
        \toprule
        Rewriting Ins (ID) & Task & Type & \multicolumn{1}{c}{Instruction Text}  \\ 
        \hline
        \multirow{4}{*}{\shortstack{Use synonyms \\ (1)}} & \multirow{2}{*}{Sentiment Understanding} & Raw & What is the caption of the image? \\
        \cline{3-4}
        ~ &  & Gen & What is the description of the image? \\
        \cline{2-4}
        
         & \multirow{2}{*}{Object Region Match} & Raw & Is the object \{text\} in \{regions\}? \{options\} \\
        \cline{3-4}
        ~ & & Gen & Is the object \{text\} located in \{regions\}? \{options\} \\
        \hline

        \multirow{10}{*}{\shortstack{Rearranging Phrases \\ (2)}} & \multirow{3}{*}{Image Caption} & Raw & Please analyze the sentiment depicted in the image and answer the question. \{question\} \{options\}  \\
        \cline{3-4}
        ~ & ~ & Gen & Please answer the question after analyzing the sentiment depicted in the image. \{question\} \{options\} \\
        \cline{2-4}
        
         & \multirow{5}{*}{Object Region Selection} & Raw & In this task, you are given the coordinates of some rectangular region in the image. You need to first localize each rectangular region and then identify what is the object in the region. The region is \{regions\}. \\
        \cline{3-4}
        ~ & & Gen & The task at hand is to localize and identify objects within a specified rectangular region in an image. The region of interest is \{regions\}. To complete the task, you will need to first localize the region and then identify the object within it. \\
        \hline
        
        \multirow{6}{*}{\shortstack{Simplyfy Language \\ (3)}} & \multirow{2}{*}{Grounded Captioning} & Raw & What is the content of \{regions\}? \\
        \cline{3-4}
        ~ & ~ & Gen & What is inside \{regions\}? \\
        \cline{2-4}
        
         & \multirow{3}{*}{Visual Attribute} & Raw & Decide which option is the attribute of the object in the given region. Region: \{regions\} \{options\}  \\
        \cline{3-4}
        ~ & & Gen & Determine the attribute of the object in the specified region: \{regions\} \{options\} \\
        \hline
        
        \multirow{5}{*}{\shortstack{Remove Redundancy \\ (4)}} & \multirow{2}{*}{Image Caption} & Raw & In this task, you will look at the image and briefly describe the image. \\
        \cline{3-4}
        ~ & ~ & Gen & Briefly describe the image you see. \\
        \cline{2-4}
         & \multirow{2}{*}{Object Grounding} & Raw & Given the description of an object, generate the region that contains this object. The description is: \{text\}  \\
        \cline{3-4}
        ~ & & Gen & Please generate the region that contains the object described by '\{text\}'. \\
        \hline
        
        \multirow{7}{*}{\shortstack{Use Active Voice \\ (5)}} & \multirow{3}{*}{Factual Checking} & Raw & Deicide if the claim can be supported by the image and the context. Context: \{context\} Claim: "\{text\}"  \{options\}   \\
        \cline{3-4}
        ~ & ~ & Gen & Evaluate the validity of the claim \{text\} \{options\} based on the image and the surrounding context \{context\} . \\
        \cline{2-4}
         & \multirow{3}{*}{Object Identification} & Raw & In this task, you are required to answer a question about the appearance of an object.\{question\} \{options\}. \\
        \cline{3-4}
        ~ & & Gen & For this task, you need to describe the appearance of an object. Please provide your answer in the format of \{question\} \{options\}. \\

        \bottomrule
        \end{tabular}
    }
    \caption{The raw and generated instructions by self-guided instruction in~\Cref{tab:meta_gen}. For better readability, we omit ``\{option\_token\}'' and simplify ''\{split/(region\_split)\_token.join(options/regions)\}'' as ``\{options/regions\}''.}
    \label{tab:instr_example}
\end{table*}

\section{\datasetname~Details}
\label{sec:mins_detail}
\subsection{Statistics of~\textsc{MIns+}}
Instance pool of~\datasetname~is same as~\textsc{MultiInstruct}, which encompasses 62 tasks in total (53 tasks in training set). 
Among the 62 tasks, 34 are predefined existing tasks while the remaining 28 are \textit{derived} from the annotations on the existing tasks, such as object matching from Visual Genome~\cite{krishna2017visual}~and non-overlapping region selection from MSCOCO~\cite{lin2014microsoft}. 

We list the statistic of~\textsc{MIns} and \datasetname~in~\Cref{tab:ds_stat}. 
Here~$\datasetname_\mathrm{MT}$ samples under 11 different temperatures (0.5 to 1.0, with an interval of 0.05). $\datasetname_\mathrm{Iter}$ only iterates twice at the default temperature 0.6, since the number of valid instructions increases exponentially (see~\Cref{tab:ds_stat}, \textsc{MIns}$\rightarrow$\datasetname$\rightarrow\datasetname_\mathrm{Iter}$). From the table we can capture a trend of increment in length as the size grows, but this is restricted to within 15\% due to the effect of length filters.

\begin{table}[ht]
    \small
    \centering
    \resizebox{\linewidth}{!}{
        \begin{tabular}{c|c|c|c|c}
        \toprule
            Dataset & \textsc{MIns} & \datasetname & $\datasetname_{\text{MT}}$ & $\datasetname_{\text{Iter}}$   \\
        \hline
        \rule{0pt}{2.5ex}
        \#Instr & 329 & 1855 & 9566 & 8724 \\
        \hline
        \rule{0pt}{2.5ex}
        Avg. Len & 14.66 & 15.46 & 16.78 & 16.50 \\
        \bottomrule
        \end{tabular}
    }
    \caption{Statistics of~\textsc{MIns}~and~\datasetname~instructions in the training set. All items under~\datasetname~are counted from filtered versions.}
    \label{tab:ds_stat}
\end{table}

\subsection{Examples of Rewritten Instructions}
We give some examples of the generated new instructions with corresponding rewriting instructions and original text  in~\Cref{tab:instr_example}.
From the table, we can find that the LLM (LLaMA-2-13b-chat) basically follows its self-generated guiding instruction to produce output.

\subsection{Length Distribution}
\begin{figure}
    \centering
    \resizebox{\linewidth}{!}{
        \includegraphics{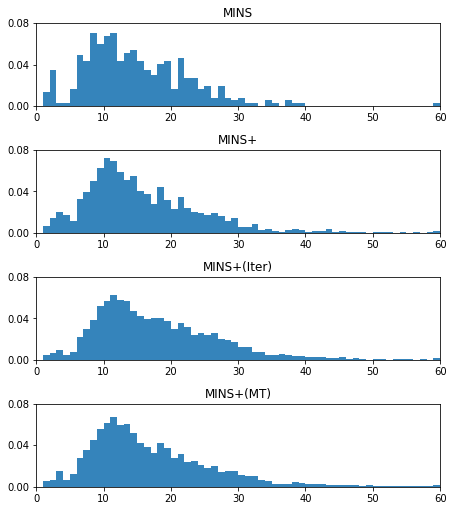}
    }
    \caption{Distributions of instruction lengths of the original and generated datasets. The horizontal axis represents the instruction length and the vertical axis is the frequency in that dataset. }
    \label{fig:ldistrib}
\end{figure}

We plot the length distribution of instructions from~\textsc{MIns}~and three~\datasetname~datasets, as shown in~\Cref{fig:ldistrib}.
Instructions whose length exceeds 60 accounts for less than 1\% in all datasets so we cut off the histogram within the range $[0,60]$.
We find that even smaller expansion version~\datasetname~can have a more smooth distribution than~\textsc{MIns}, and larger size~\datasetname~($\datasetname_\textrm{Iter}$~and~$\datasetname_\textrm{MT}$) continue to flat the length distribution histogram.

\subsection{Wording Characteristics}
\begin{figure*}[ht]
    \centering
    \begin{subfigure}[c]{\textwidth}
        \includegraphics[width=\textwidth,trim=0cm -0.7cm 0cm 0]{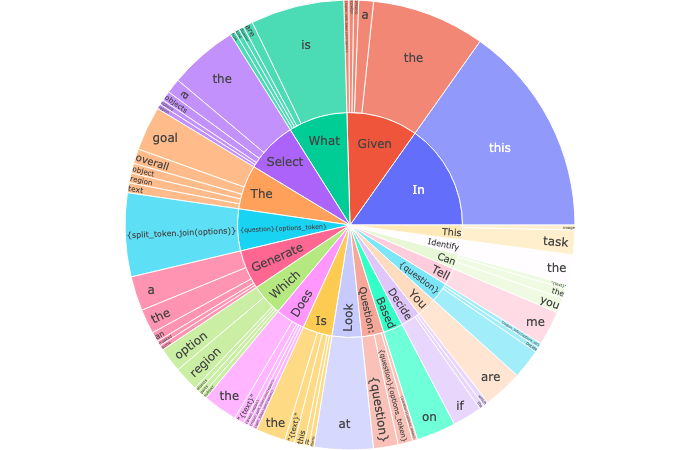}
        \caption{The 20 most common first words along with their 5 most common next words in all the instructions from~\textsc{MultiInstruct}~training set.}
        \label{fig:f2words_raw}
        \vspace{1em}
    \end{subfigure}
    \begin{subfigure}[c]{\textwidth}
        \includegraphics[width=\textwidth, trim=0cm -0.7cm 0cm 0]{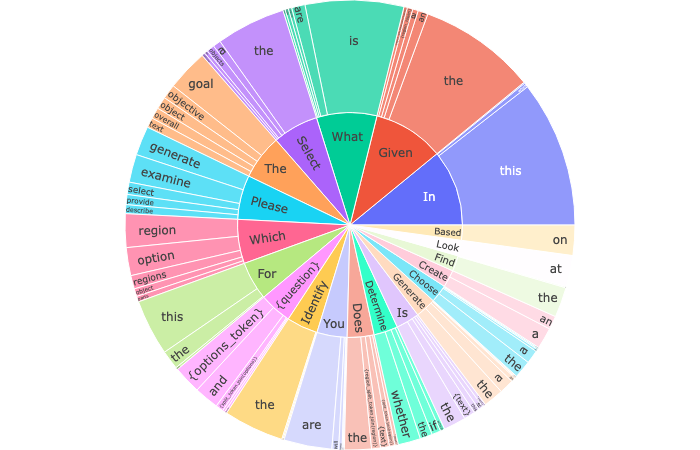}
        \caption{The 20 most common first words along with their 5 most common next words in all the instructions from~\datasetname.}
        \label{fig:f2words_plus}
    \end{subfigure}
    \caption{(Figure 1/2) Instruction prefix (including placeholders enclosed by brackets) distribution in (a)~\textsc{MIns} (b)~\datasetname.}
    \label{fig:words_distrib1}
\end{figure*}

\begin{figure*}[ht]
    \centering
    \begin{subfigure}[c]{\textwidth}
        \includegraphics[width=\textwidth,trim=0cm -0.7cm 0cm 0]{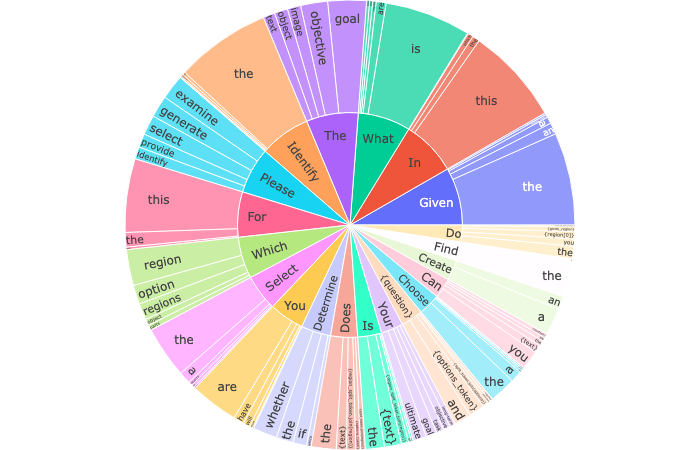}
        \caption{The 20 most common first words along with their 5 most common next words in all the instructions from~\datasetitername~training set.}
        \label{fig:f2words_plusiter}
        \vspace{1em}
    \end{subfigure}
    \begin{subfigure}[c]{\textwidth}
        \includegraphics[width=\textwidth, trim=0cm -0.7cm 0cm 0]{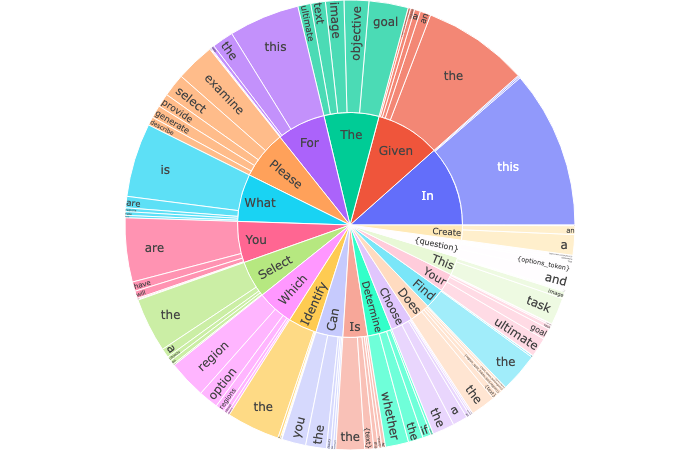}
        \caption{The 20 most common first words along with their 5 most common next words in all the instructions from~\datasetmtname.}
        \label{fig:f2words_plusmt}
    \end{subfigure}
    \caption{(Figure 2/2) Instruction prefix (including placeholders enclosed by brackets) distribution in (a)~\datasetitername~(b)~\datasetmtname.}
    \label{fig:words_distrib2}
\end{figure*}

To observe the wording diversity of~\textsc{MIns+} and comparing the shift with~\textsc{MIns}, 
we further plot the distribution of instruction wording---the first two words of highest frequency in~\Cref{fig:words_distrib1}~and~\Cref{fig:words_distrib2}.
These figures explicitly reveal a shift from raw wording frequency distribution to generated wording frequency distribution, which appears in
\begin{itemize}
    \item Change in the area of sectors which represent the relative frequency in the overall dataset. For example, the area of ``In'' and ``Given'' diminish. 
    Some pairs of words even swap the relative positions that imply the frequency ranking (frequency descends counter-clockwisely), such as ``The'' and ``What'' which respectively rank 3rd and 5th in~\textsc{MIns}~, but rank 5th and 3rd in~\datasetname. 
    \item New sentence patterns created in~\datasetname, such as those which begin with ``For'', ``You'' and ``Please'', etc.
\end{itemize}

\section{Training and Evaluation Details}
\subsection{Hardware and Hyperparameters}
\label{sec:hyper}
We tune OFA on 4 A6000 GPU of 46G RAM and IBLIP on 2 A100 GPU of 80G RAM.
We follow most of the default settings in each tuning version except the batch size for gradient accumulation. 
The hyperparameter setting for both models can be found in~\Cref{tab:hyper}.

\begin{table}[ht]
    \centering
    \resizebox{\linewidth}{!}{
        \begin{tabular}{c|c|c}
        \toprule
           Model & OFA & IBLIP \\
        \hline
           Learning Rate  & 1e-5 & 1e-5 \\
           Optimizer & Adam & AdamW \\
           Adam $\beta$ & 0.9, 0.999 & 0.9, 0.999 \\
           Epoch & 8 & 8 \\
           Batch Size & 64 & 128 \\
           Gradient Accumulation & 4 & 8 \\
           Image Size & 480$\times$480 & 224$\times$224 \\
           FP16 & True & True \\
        \bottomrule
        \end{tabular}
    }
    \caption{Hyperparameters for tuning on each model.}
    \label{tab:hyper}
\end{table}

\subsection{Tuning Details}
We fine-tune the targeted models on~\datasetname~for 5 epochs, utilizing the default hyperparameters settings. For the 1K/10K (per task) version, the total instance pool size is 59K/564K (with some tasks having fewer than 10K samples). We validate the checkpoint of each epoch for the 59K version and every 2K steps for the 564K version.

\subsection{More Details of Evaluation Tasks}
Generally, the tasks can be categorized into three types: 1) Generative task, including Visual Text Extraction (VTE)~\cite{kiela2020hateful}, TextVQA~\cite{singh2019towards}, Visual Dialogue (VisDial) and Commonsense VQA (CVQA); 2) Multiple-choice task, which provides instance-dependent options for each question, including several VQA tasks (Grounded VQA~\cite{zhu2016visual7w}, Science QA~\cite{lu2022learn}, IconQA~\cite{lu2021iconqa}, VizWiz~\cite{gurari2018vizwiz}) plus Disaster Classification (DC)~\cite{alam2023medic}; 3) Discriminative tasks, including Visual Entailment, Visual Spatial Reasoning and Natural Language for Visual Reasoning, which can be viewed as a special variant of multiple-choice or \textit{multiple-choice with fixed options}. 
For example, the options are restricted within ``yes/no/not sure'' in VE and ``yes/no'' in other tasks.
Following the evaluation process in these papers, we report Rouge-L~\cite{lin2004rouge}~score for generative tasks and accuracy for the other two types of tasks.

For each test task, we also determine whether the instruction tuning set includes the same task (\textbf{task overlap}) or if the test dataset overlaps with both the tuning and pre-training datasets (\textbf{data overlap}), based on the visible data sources reported in the corresponding dataset papers~\cite{xu2022multiinstruct, dai2305instructblip}, as illustrated in~\Cref{tab:eval_tasks}.
This could help us make an in-depth analysis. 
Regarding each base model, we only report the~\textbf{held-out}~tasks for fair comparison. 
To this end, we exclude the results of VTE and VisDial on IBLIP in~\Cref{tab:res_mins}. 

\begin{table}[ht]
    \centering
    \resizebox{0.97\linewidth}{!}{
    \begin{tabular}{c|c|c|c|c|c|c}
    \toprule
         \multirow{2}{*}{Source} & \multirow{2}{*}{Task Name} & \multirow{2}{*}{Type} & \multicolumn{2}{c|}{\textsc{MIns}} & \multicolumn{2}{c}{IBLIP} \\
        \cline{4-7}
         ~ & ~ & & T & D & T & D \\
        \hline
        \multirow{9}{*}{~\textsc{MIns}} & VTE & G & 
        & & $\checkmark$ & $\checkmark$ \\
        ~ &TextVQA & G & $\checkmark$  &  & $\checkmark$  &  \\
        ~ &VisDial & G & $\checkmark$ &  & $\checkmark$ & $\checkmark$ \\
        ~ &CVQA & MC &  & & & \\
        ~ &DC & MC &  & & & \\
        ~ &Grounded VQA & MC & $\checkmark$ & & & \\
        ~ &VE & D & & & & \\
        ~ &VSR & D & & & & \\
        ~ &NLVR & D & & & & \\
        \hline
        \multirow{3}{*}{\shortstack{IBLIP-Bench}} & ScienceQA & MC & & & $\checkmark$ & \\
        ~ & IconQA & MC & & & & \\
        ~ & VizWiz & MC & & & $\checkmark$ & \\
    \bottomrule
    \end{tabular}
    }
    \caption{Evaluation tasks and their relationship with pretraining and training datasets. G: generative task, MC: multiple-choice task, D: discriminative task. T and D respectively denote task overlap and data overlap.}
    \label{tab:eval_tasks}
\end{table}

\subsection{Evaluation Methods}
\label{sec:eval_method}
We describe the evaluation formulation for all the tasks and models respectively. 
The tasks can be generally divided into 2 types: generation tasks and classification tasks (including multiple-choice tasks).

\paragraph{Generation Tasks} The evaluation for generation tasks on both models are same and pretty straightforward. We let the model generate responses under fixed hyperparameters (e.g., beam size, maximum and minimum length) and compute the ROUGE-L score with ground truths.

\paragraph{Classification Tasks} Evaluation for classification tasks is quite different using the two models. For OFA, we still formulate as generation tasks and the accuracy is defined as:
\begin{equation}
    Acc = \frac{1}{N}\sum_i \mathbf{1}(pred_i=truth_i)
\end{equation}

For IBLIP, we calculate the sequence-to-sequence CE loss from prompt to the option-appended prompt. If an instance has $O$ options, then the criterion is: the prediction is correct \textbf{if the truth-appended prompt earns the minimum loss among all options}, 
\begin{equation}
    Acc = \frac{1}{N}\sum_i \mathbf{1}(\arg\min_{j}(CE(in_i,out_{i,j})) = idx(a_i))
\end{equation}
where $j\in\{1,2,...,\vert O \vert\}$ and $idx(a)$ is the index of truth (answer) in the given option list.

\subsection{Templates for IBLIP-Bench Tasks}
\label{sec:eval_templates}
We provide the evaluation templates for three IBLIP-Bench tasks (Science QA, Icon QA, VizWiz) in~\Cref{tab:eval_template}.
The raw instruction can be found in~\cite{dai2305instructblip}, which we show in the row ``IBLIP''. 
The instructions in the ``OFA'' row are what we create from the raw instruction by modifying delimiters and special tokens to be compatible with OFA's expression style.
Here to simplify the expression, we assume there are two options, ``Apple'' and ``Pear'' and the correct answer is always ``Apple''. 
We also list the target strings to judge whether the output is correct.

\begin{table*}
    \resizebox{\linewidth}{!}{
        \begin{tabular}{c|c|c|c}
        \toprule
           Task & Style & Template & Target  \\
        \hline
           \multirow{2}{*}{Science QA}  & IBLIP & Context: \{C\} Question: \{Q\} Options: Apple||||Pear. Answer: & Apple \\
           \cline{2-4}
           ~ & OFA & \{C\} Question: \{Q\} [Options]: Apple||||Pear & Apple \\
        \hline
           \multirow{2}{*}{Icon QA}  & IBLIP & Question: \{Q\} Options: Apple||||Pear. Short answer: & Apple \\
           \cline{2-4}
           ~ & OFA & \{Q\} [Options]: Apple||||Pear & Apple \\
        \hline
        \end{tabular}
    }
    \caption{Templates used for evaluation on corresponding models. ``\{Q\}'' and ``\{C\}'' respectively represent text in the question and context field. The main differences concentrate on delimiters and generation prefix.}
    \label{tab:eval_template}
\end{table*}

\section{Result Visualization}
In this section, we visualize and compare the results tuned on~\textsc{MIns}~and~\textsc{MIns+}~series~ in~\Cref{tab:visual_example_vte}~and~\Cref{tab:visual_example_dc}.
In both cases, the MLLM either misunderstands the target of the task or fails to predict the correct answer.

\begin{table*}[ht!]
  \begin{minipage}{\textwidth}
    \centering
    \scalebox{0.95}{
    \begin{tabular}{l p{12.5cm} }
    \toprule
     \multicolumn{2}{l}{\bf Evaluation result example 1 (Generation): Visual Text Extraction}  \\
    \midrule
    Input Image &  \includegraphics[align=c,height=4.5cm]{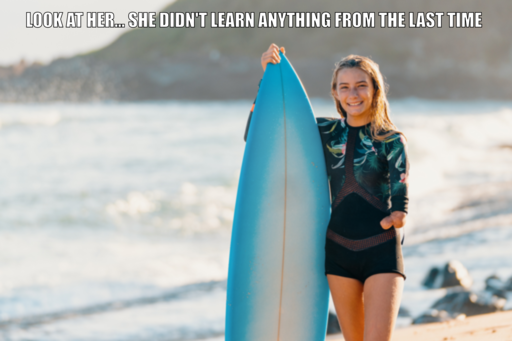} \\
    Target & look at her... she didn't learn anything from the last time \\
    \midrule
    \textbf{Instruction 1} & What is the text written on the image? \\[0.6cm]
    \textsc{MIns} (59K) &  a woman holding a surfboard \\[5pt]
    \datasetname~(59K) & look at her, she didn't learn anything from the last time \\
    \midrule
    \textbf{Instruction 2} & This image contains some text. For this task, you need to look at the image carefully and identify all the text in the image. The text in the image is \\[1cm]
    \textsc{MIns}~(564K) &  yes \\[5pt]
    \datasetitername~(564K) & she didn't learn anything from the last time \\
    \bottomrule
    \end{tabular}
    }
    \vspace{2mm}
    \captionof{table}{Response comparison between results by OFA tuned on different versions of~\textsc{MIns}. In both cases on~\textsc{MIns}~fails to understand the question and produces weird output. The incorrect response seems as captioning towards instruction 1 and yes/no question towards instruction 2.}
    \label{tab:visual_example_vte}  
    \end{minipage}
\end{table*}

\begin{table*}[ht!]
  \begin{minipage}{\textwidth}
    \centering
    \scalebox{0.95}{
    \begin{tabular}{l p{12.5cm} }
    \toprule
     \multicolumn{2}{l}{\bf Evaluation result example 2 (Multiple Choice): Disaster Classification}  \\
    \midrule
    Input Image &  \includegraphics[align=c,height=4.5cm]{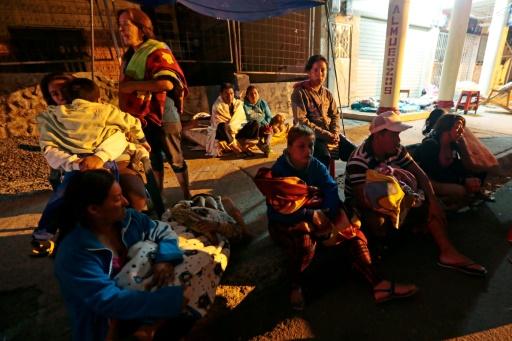} \\
    Target & not disaster \\
    \midrule
    \textbf{Instruction 1} & What disaster happens in the image?

    [Options]: earthquake||||informative||||not disaster||||mild damage||||other disaster||||hurricane||||little or none damage||||not informative||||severe damage||||landslide||||flood||||fire \\[2cm]
    \textsc{MIns} (59K) & earthquake  \\[5pt]
    \datasetname~(59K) & not disaster \\
    \midrule
    \textbf{Instruction 2} & According to the image, what kind of disaster happened? Choose the correct answer from options.
    
    [Options]: earthquake||||informative||||not disaster||||mild damage||||other disaster||||hurricane||||little or none damage||||not informative||||severe damage||||landslide||||flood||||fire \\[2.6cm]
    \textsc{MIns}~(564K) &  hurricane \\[5pt]
    \datasetmtname~(564K) & not disaster \\
    \bottomrule
    \end{tabular}
    }
    \vspace{2mm}
    \captionof{table}{Response comparison between results by OFA tuned on different versions of~\textsc{MIns}. In this example, both baselines successfully perceive the task type and select from options as output, but they fail to predict correctly.}
    \label{tab:visual_example_dc}  
    \end{minipage}
\end{table*}

\end{document}